# The active visual sensing methods for robotic welding: review, tutorial and prospect

ZhenZhou Wang*
School of Computer Science and Technology,
Huaibei Normal University, Huaibei City, China.
(*e-mail: 18553358239@163.com).

*Abstract*—**The visual sensing system is one of the most important parts of the welding robots to realize intelligent and autonomous welding. The active visual sensing methods have been widely adopted in robotic welding because of their higher accuracies compared to the passive visual sensing methods. In this paper, we give a comprehensive review of the active visual sensing methods for robotic welding. According to their uses, we divide the state-of-the-art active visual sensing methods into four categories: seam tracking, weld bead defect detection, 3D weld pool geometry measurement and welding path planning. Firstly, we review the principles of these active visual sensing methods. Then, we give a tutorial of the 3D calibration methods for the active visual sensing systems used in intelligent welding robots to fill the gaps in the related fields. At last, we compare the reviewed active visual sensing methods and give the prospects based on their advantages and disadvantages.**

## I. INTRODUCTION

WELDING is one of the most fundamental processing means of industrial manufacturing, but the welding process will produce smoke, arcs and metal splashes, etc., resulting in the bad welding environment. As a result, traditional welding enterprises face the problems of low production capacity, recruitment difficulties and low profits, and the use of welding robots can solve these problems well. After a long period of development, the welding robot has developed from the "automatic welding special equipment" to the "teaching playback" mode robot at the present stage [1]. For welding parts with complex geometry and weld seams, the operator needs to do complex teaching and programming work on the welding path planning of the welding robot, and the quality of the welding depends largely on the operator's experience. In order to avoid complex teaching playback mode, it is necessary to develop intelligent welding robots. In order to realize intelligence, it is necessary for welding robots to have strong perception ability, and visual sensing technology, as the most effective perception means, has become the key to achieve intelligent welding robots. The welding robot can self-learn and think through the visual sensing technology, so as to realize intelligence. Specifically, visual sensing technology can overcome the influence of various uncertainties on welding quality in the welding process, so in the past few decades, researchers have proposed a large number of visual sensing methods to monitor the welding process. Overall, the visual sensing methods for robotic welding could be divided into two categories: (1) the passive visual sensing methods and (2) the active visual sensing methods.

The passive sensing methods use the cameras only to obtain the images of the welding targets directly for welding path planning [2], seam tracking [3], metal transfer monitoring [4], weld bead defect detection [5-7] and weld penetration monitoring [8]. Because the obtained welding images may vary greatly according to the welding processes, they usually require specific image processing algorithms to extract the useful information. In addition, the lack of texture or color and the interference during the welding will cause the accuracies of the passive sensing methods to be very low. Therefore, the active visual sensing methods are more widely applied than the passive visual sensing methods [1].

The active visual sensing methods combine the structured light (SL) patterns with one or two cameras to obtain the 3D information of the welding targets. Because the SL patterns add texture or color to the welding images and SL are usually strong enough to resist the interference during the welding, the accuracies of the active visual sensing methods are usually high enough to measure the 3D information of the welding targets robustly. Thus, the active visual sensing methods become the best choices for the perception systems of the welding robots. In this paper, we review the active visual sensing methods that were used for robotic welding in the past decades. We divide these active visual sensing methods into four categories: (1), the active visual sensing methods for seam tracking; (2), the active visual sensing methods for weld bead defect detection; (3), the active visual sensing methods for 3D weld pool geometry measurement; (4), the active visual sensing methods for welding path planning.

The active visual sensing methods for seam tracking project the laser line SL on a small region of the weldment ahead of the welding torch during welding for real-time seam identification and tracking [9-32]. The purpose of seam tracking is to correct the position disturbance of the weld seam under the torch caused by weldment deformation, heat conduction and gap changes. If the measured seam point is too close to the welding torch, it will be affected greatly by the strong arc light and splash noise. Consequently, the seam point extraction accuracy will be decreased severely. If the measured seam point is too far away from the welding torch, seam tracking will be meaningless since the objective of seam tracking is to rectify the position disturbance of the seam under the welding torch caused by weldment distortion, heat spread and gap variability. Therefore, the tracking position of the weld seam needs to be selected carefully. Unfortunately, there are no research work about selecting the optimal tracking position yet. In recent years, the research of seam tracking was mainly focused on how to accurately extract the seam point from the noisy images. For instance, different deep-learning-based methods have been proposed to compute the weld seam tracking points more robustly under strong arc light [16-19]. However, seam tracking has still remained challenging for the multi-layer/multi-pass welding so far [9-13].

The active visual sensing methods for welding bead defect detection project the laser line SL on a small region of the weldment behind the welding torch during welding for in situ welding bead defect detection [33-36] or project the SL patterns on the whole weldment after welding for offline welding bead defect detection [37-40]. In recent years, the research of weld bead defect detection was mainly focused on increasing the automation level. To increase the automation level of the online weld bead inspection system, researchers came up with algorithms to determine the weld bead type automatically [33], came up with algorithms to recognize the type of the weld defect automatically [36] and proposed algorithms to move the laser line SL sensors automatically along the weld seam for offline weld bead defect detection [38-39]. Researchers also changed the laser line pattern to the sinusoidal plane pattern to increase the efficiency of offline weld bead defect detection [40].

The active visual sensing methods for 3D weld pool geometry measurement project the laser SL on a small region of the weldment beneath the welding torch during welding for in situ 3D weld pool geometry measurement and penetration control [41]. The weld pool geometry determines or affects the weld appearance and the generation of weld defects. Hence, the skilled welders always utilize the observed weld pool information to control its geometry and to get a good weld quality [42]. Weld pool monitoring is extremely difficult because the external welding environment is very harsh. In addition, the surface of the molten weld pool is mirror surface. The reflection from mirror surfaces is view-angle dependent, which adds the ambiguity of slope and height to the traditional SL or active stereo vison (ASV) methods. To address these challenges, researchers have come up with many different active sensing methods that include the iterative-estimation based methods [43-49] and the ray-intersection based methods [50-51]. The iterative-estimation based methods match the projected SL pattern with the designed pattern and then estimate the three-dimensional shape of the weld pool iteratively based on the governing reflection law [43]. These methods suffer from the ambiguity of slope and height. Consequently, an analytical solution does not exist for these methods. To improve the robustness, researchers combined a single laser line 3D scanner and a passive camera to reconstruct of the GMAW weld pool [49]. However, the analytical solution of the 3D weld pool geometry still was not achieved. So far, only the ray-intersection based methods could measure the 3D shape of the weld pool with an analytical solution robustly [50-51].

The active visual sensing methods [52-79] for welding path planning project the SL patterns on the weldment before or during welding for 3D weld seam reconstruction. The reconstructed 3D weld seam point clouds could be used for the guidance of many robotic welding applications, such as multilayer positioning [52] and offline programming [53] to overcome the drawbacks of traditional welding robots. The currently widely used "teaching and playback " mode and off-line programming mode requires the operator to carry out the complicated teaching and programming work on the welding path planning for the welding robot. The operator needs to visually inspect the weld seam at first and then select the teaching points or the starting point for programming. Consequently, the welding quality are highly dependent on the operator's experience. The automatic welding path planning based on the point cloud of the weld seam could simplify the complicated teaching and programming work. In addition, automatic welding path planning could also improve the welding quality, efficiency and automation level of the welding robot. Currently, the active sensing methods for welding path planning include the single laser line based SL methods [54-63], the phase shifting profilometry (PSP) based SL methods [64-65], the multiple line based ASV methods [66-72], the Fourier transform profilometry (FTP) based active stereo vision methods [73] and the laser speckle based ASV methods [74-79].

## II. THE ACTIVE VISUAL SENSING METHODS FOR SEAM TRACKING

At present, most of the used welding robots belong to the teach-and-playback robots and they have some fatal weaknesses. (1), they require a great amount of time to be taught in advance, which leads to low efficiency. (2), they cannot rectify deviations on-line during the welding process. However, the deviation of the weld droplet from the ideal weld seam position will cause poor weld quality. With the seam tracking system, the actual weld pass position could be recognized to correct any deviations on line and also to perform reasonable, but limited path planning for the welding robots.

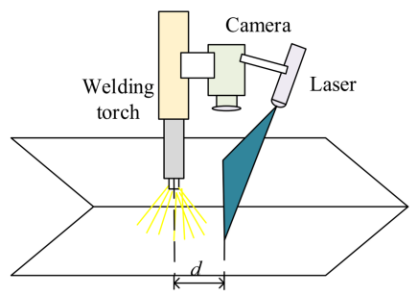

Fig. 1. The schematic diagram of a weld seam tracking system from [9].

The schematic diagram of a typical weld seam tracking system from [9] is shown in Fig. 1. The single straight laser line SL pattern is projected onto the weldment ahead of the welding torch along its moving direction. The laser line image is acquired by a camera and then processed in real-time to extract the seam tracking point. The single straight laser line as demonstrated in Fig. 1 is the most frequently used SL pattern for seam tracking because of its simplicity and effectiveness in computing the feature points.

According to the welding demands, the welding processes could be divided into two categories: (1), the multi-layer and multi-pass welding process [9-13] and (2), the single-pass welding process [14-32]. Both welding processes require seam tracking to improve the welding quality. In general, it is much more challenging to compute the seam points for the multi-pass welding processes than to compute the seam points for the single-pass welding processes.

### A. Multi-pass weld seam tracking

The multi-pass welding processes usually produce multi-layers of weld seams and thus they are also called the multi-layer and multi-pass welding or additive welding. With the increasing number of layers, weld seam tracking becomes progressively more challenging. The main reason is that the characteristics of high-rise weld seams are less apparent, which will result in the accumulation of errors from layer to layer and the significant deviation from the ideal state. A typical multi-layer and multi-pass welding process is used in [9]. There are six passes in total and they generate three layers in total. The first pass generates the first layer and the second pass and the third pass generate the second layer. The fourth pass, the fifth pass and the sixth pass generate the third layer. As it turned out by the experimental results, it is easy to determine the seam point from the feature points for the first pass. However, when the number of the pass increases, the difficulty of determining the seam point increases accordingly because the characteristics of the weld center line is becoming more and more not obvious with the increase of weld layer. Though the efficient residual factorized ConvNet (ERFNet) proposed in [9] has good segmentation results regardless where the laser stripe is located and where the weld features are located in the image, it may be difficult to select the right feature from all the extracted features in the higher weld layers.

Another typical multi-layer and multi-pass welding process was used in [10] and the pixel-wise spatial pyramid network (PSPNet) was proposed to extract the center line of the laser stripe and the weld seam point. Different from the deep-learning based method proposed in [9], the PSPNet outputted the weld seam point directly instead of all the feature points. Their experimental results showed that the weld seam point of the multi-layer and multi-pass welding process could be tracked robustly by the deep-learning based methods. Compared to the traditional methods, such as the visual attention model [11], the deep learning based seam tracking methods for the multi-layer and multi-pass welding processes have made significant progress.

### B. Single-pass weld seam tracking

For the single-pass weld seam tracking, it is easy to determine the seam point because it usually is the intersection point of two straight lines. On the other hand, straight line detection is well-studied problem. When the noise is not severe, the seam point can be computed robustly [14]. When the arc light is strong but the laser line is far enough from the torch, the seam point can also be computed robustly [15]. When the projected laser line is covered by the strong arc light, the seam point could only be estimated with uncertainty [16].

Besides the widely used single straight laser line [9-28], there are also other types of SL patterns used by some researchers for the single-pass weld seam tracking [29-32]. For instance, the authors in [29-30] used three parallel straight laser lines for seam tracking and the center laser line alone is used to compute the seam point. The straight laser lines on the two sides are used to enhance features and constrain the searching region.

As a matter of fact, if all the three laser lines are used to calculate the seam points, they could do a better path planning job for the curved seams. The authors in [31] used the cross straight laser lines for seam tracking. The cross mark was first detected and then used to constrain the search range/direction for the seam point detection. The authors in [32] used the circular laser for seam tracking and two feature points were computed for seam tracking.

### C. Multi-pass weld seam tracking

Compared to the single-pass welding process, the multi-pass welding processes are much more complex. The multi-pass welding processes usually produce multi-layers of weld seams and thus they are also called the multi-layer and multi-pass welding or additive welding. With the increasing number of layers, weld seam tracking becomes progressively more challenging. The main reason is that the characteristics of high-rise weld seams are less apparent, which will result in the accumulation of errors from layer to layer and the significant deviation from the ideal state. A typical multi-layer and multi-pass welding process is used in [9]. There are six passes in total and they generate three layers in total. The first pass generates the first layer and the second pass and the third pass generate the second layer. The fourth pass, the fifth pass and the sixth pass generate the third layer. As it turned out by the experimental results, it is easy to determine the seam point from the feature points for the first pass. However, when the number of the pass increases, the difficulty of determining the seam point increases accordingly because the characteristics of the weld center line is becoming more and more not obvious with the increase of weld layer. Though the efficient residual factorized ConvNet (ERFNet) proposed in [9] has good segmentation results regardless where the laser stripe is located and where the weld features are located in the image, it may be difficult to select the right feature from all the extracted features in the higher weld layers.

Another typical multi-layer and multi-pass welding process was used in [10] and the pixel-wise spatial pyramid network (PSPNet) was proposed to extract the center line of the laser stripe and the weld seam point. Different from the deep-learning method in [9], the PSPNet outputted the weld seam point directly instead of all the feature points. Their experimental results showed that the weld seam point of the multi-layer and multi-pass welding process could be tracked robustly by the proposed method. Compared to the traditional methods, such as the visual attention model [11], the deep learning based seam tracking methods for the multi-layer and multi-pass welding processes have made significant progress.

## III. THE ACTIVE VISUAL SENSING METHODS FOR WELD BEAD DEFECT DETECTION

### A. In situ weld bead defect detection

The schematic diagram of the in-situ weld bead defect detection system [33-36] is similar to that of the seam tracking system shown in Fig. 1. The major difference is that the laser line is projected onto the weldment behind the welding torch instead of ahead of the welding torch along its moving direction. The laser line image is acquired by a camera and then processed in real-time to detect the weld bead defects. The weld bead defects include the bead misalignment, displacement of weldment, reinforcement height mutation and undercut. Fig. 2 illustrates the terms of weld bead defects. The turning points, $p_1 = (x_1, y_1)$, $p_2 = (x_2, y_2)$, $p_3 = (x_3, y_3)$ and $p_4 = (x_4, y_4)$ are computed by image processing and camera calibration.

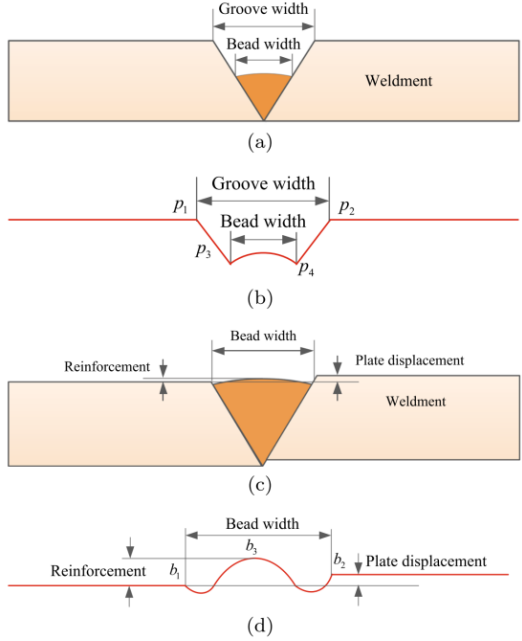

Fig. 2. Illustration of the weld bead defects in multi-layer weld [33]: (a) Cross section of the weld bead in root-pass weld. (b) Laser line profile in the root-pass weld. (c) Cross section of the weld bead in cap weld. (d) Laser line profile in the cap weld.

Weld bead misalignment defect can be detected based on the lateral symmetry of the weld groove and the trapezoidal formed by four turning points will be asymmetrical when it occurs. When the plate displacement is larger than an acceptable value, the displacement defect occurs and it can be detected based on the Hough transform of the laser line profiles. Reinforcement height is defined as the maximum distance between points on the weld bead profile and the weld baseline. When the reinforcement height of weld bead changes greatly and abruptly, reinforcement height mutation defect occurs. Undercut defect occurs when the weld bead width is greater than the groove width. It can be detected by comparing the distance $W_b$ between the two border points $b_1 = (x_1, y_1)$ and $b_2 = (x_2, y_2)$ with the average groove width $W_g$. That is to say, if $W_b > W_g$, undercut defect occurs.

Similar to weld seam tracking, the most challenging part of weld bead defect detection is to robustly extract the feature points, $p_1$, $p_2$, $p_3$, $p_4$, $b_1, b_2$ and $b_3$ that are demonstrated in Fig. 2. The feature points were extracted by the second-order difference of the y-coordinate of the center point of the laser line in [33]. However, it is very challenging to select the feature point $p_4$ among quite a few valley points robustly and automatically because we could not know in advance which valley point corresponds to $p_4$. Therefore, a more robust feature extraction method is required in these applications.

### B. Offline weld bead defect detection

The offline weld bead defect detection system in [37] uses the same laser scanner as the in-situ system. However, the in-situ weld bead defect detection system could stop the welding process immediately to save cost when severe defects occur while the offline system could not. Thus, the in-situ system is more desirable for robotic welding. A more efficient offline weld bead defect detection system based on the phase shifting profilometry (PSP) sinusoidal patterns with four phase shifts was proposed in [40]. It projected the sinusoidal patterns on the weldment and reconstructs its 3D shape as a whole. Then, the weld bead defects could be detected at each two-dimensional slice along the weld seam direction. Although its efficiency has been increased greatly compared to the single straight laser-line-pattern based system [37], the PSP based system is still suffering the following drawbacks. Firstly, the phase shifting based sinusoidal patterns are more susceptible to the influence of the external light and the surface reflectivity of the weldment compared to the laser line patterns. Secondly, it usually requires to project more than ten phase shifting sinusoidal patterns to reconstruct the 3D shape robustly. For instance, the designed 3D scanner in [40] is three carrier-frequency four-step PSP and it needs to project 12 phase shifting sinusoidal patterns to reconstruct the 3D weldment, which is less efficient compared to the single-shot 3D reconstruction methods. The principle of PSP will be described in the fifth section.

## IV. THE ACTIVE VISUAL SENSING METHODS FOR WELD POOL 3D GEOMETRY MEASUREMENT

It is difficult to ensure good penetration in robotic welding. Visual sensing is one of the most prosperous ways for penetration sensing and it could acquire abundant information as a welder's eyes. However, the passive sensing methods could not avoid the effect of the strong arc light and the extracted useful information may be decreased greatly. The active sensing methods utilize the strong laser dot or line to offset the effect of arc light. Thus, it could obtain more useful information to measure the 3D geometry of the molten weld pool.

### A. Iterative-estimation based weld pool 3D geometry measurement

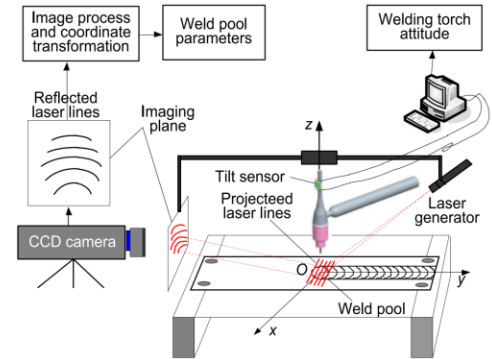

Fig. 3. The schematic diagram of a typical iterative-estimation based weld pool measurement system from [46].

The schematic diagram of a typical iterative-estimation based weld pool measurement system from [46] is shown in Fig. 3. The multiple-straight-laser-line SL pattern is projected onto the molten weld pool below the welding torch along its moving direction. The reflected laser line image is acquired by a camera and then processed with specific designed image processing algorithms [47-48] to obtain the laser lines robustly. The reflected laser lines are usually very fuzzy and the grayscale of the background is not uniformly distributed, which makes the laser line segmentation very challenging. The same laser line may break into several parts and different laser lines may interlace or overlap, which makes the laser line clustering very challenging.

Besides the multiple-straight-laser-line SL patterns, the laser dot matrix SL patterns are also frequently used in weld pool measurement [43-45, 50-51]. Compared to the dot pattern, the line pattern has the advantages of higher resolution, higher segmentation accuracy and better clustering accuracy. The dot-pattern based weld pool measurement methods could be converted to the line-pattern based weld pool measurement methods with proper modifications and vice versa. The measurement of the 3D weld pool geometry is an inverse problem of the reflection law and an analytical solution does not exist. To resolve this issue, an iterative method is needed to find an optimally estimated three-dimensional geometry [43]. The principle of the iterative-estimation based weld pool measurement methods could be summarized as follows.

*Step1:* The projected dot or line is extracted from the acquired image and then matched with the designed pattern.

*Step2:* The three-dimensional weld pool geometry from these matched discrete point sets is estimated iteratively based on the reflection law.

### B. Ray-intersection based weld pool 3D geometry measurement

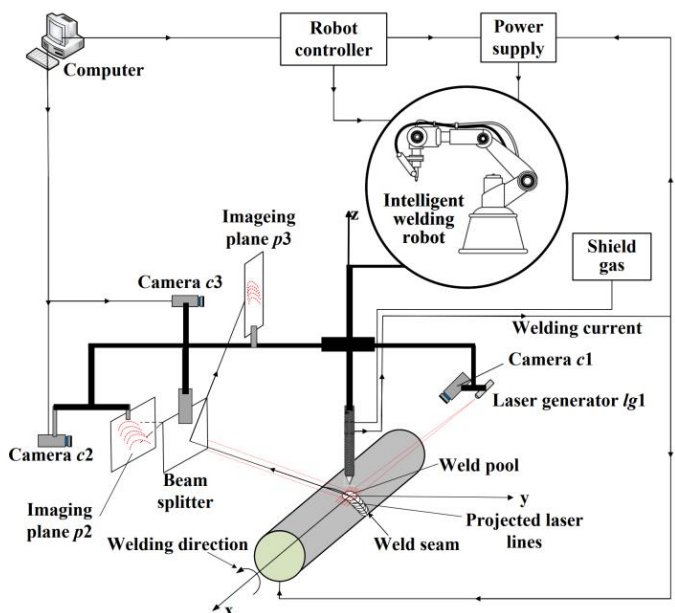

Fig. 4. The schematic diagram of the ray-intersection-based weld pool measurement system [50].

The schematic diagram of the ray-intersection based weld pool measurement system from [50] is demonstrated in Fig. 4. The laser-dot SL pattern is projected by a SNF laser onto the molten weld pool below the welding torch along the welding direction of the pipe weldment. The laser dots are reflected by the weld pool onto a beam splitter at first. The beam spitter splits the reflected rays into two parts equally. Half rays pass through the beam spitter and imaged on the imaging plane $p2$. Half rays are reflected by the beam spitter and imaged on the imaging plane $p3$. Camera $c1$ is used to compute the equations of the incident rays in the world coordinate system. Camera $c2$ is used to compute the world coordinates of the imaged laser dots on $p2$ and Camera $c3$ is used to compute the world coordinates of the imaged laser dots on $p3$. The principle of the ray-intersection based weld pool measurement method is as follows.

*Step1:* With two intersection points on $p2$ and $p3$ respectively in the 3D world coordinate system, the equation of each reflected ray is determined.

*Step2:* With the known equations of the incident ray and its corresponding reflected ray, the intersection point on the surface of the mirror reflection weld pool surface is then determined analytically.

## V. THE ACTIVE VISUAL SENSING METHODS FOR WELDING PATH PLANNING

As reviewed and described in section 2, seam tracking based on the single straight laser line could not ensure good tracking results for the multi-layer and multi-pass welding processes. In such situations, multi-layer weld seam positioning and welding path planning become important in enhancing the tracking accuracy [52]. In addition, when the geometries of the weldments and the weld seams are complex, the accuracies of seam tracking will be low. Automatic welding path planning based on the measured 3D weld seam not only could simplify the complicated teaching and programming work, but also could improve the welding quality, efficiency and automation level [53]. Accordingly, the sensing methods for welding path planning are of great importance. Compared to the passive sensing methods, the active sensing methods are much more accurate [80]. Thus, they have been frequently used to measure the 3D point cloud of the weld seam for welding path planning in recent years [54-79].

### A. Structured light methods based on single laser line

The schematic diagram of the weld seam 3D measurement system based on the single laser line SL [54-63] is similar to that of the seam tracking system shown in Fig. 1. The similarity lies in that the laser line is projected onto the weldment ahead of the welding torch along its moving direction. The difference lies in that the laser line needs to be moved several times to measure the weld seam offline in advance before the welding begins. For instance, the weld seam 3D measurement system from [54] measured the weld grove for a tubular T-joint by moving the single laser line along the $x$ axis of the world coordinate system with the step size of 20 mm at each measurement. All the measured 3D curves form the 3D point cloud of the weldment. Then specific image processing algorithms need to be designed to extract the weld seam for welding path planning.

Initial welding point measurement is a challenging task of welding path planning and the single laser line SL has been frequently used for initial welding point measurement in recent years [61-63]. The principle of the initial welding point measurement methods could be summarized as follows.

*Step1:* The laser line is projected on the weldment at a relatively far position ahead of the welding torch. Then, the welding robot moves the laser line along the opposite direction of welding and the detection starts.

*Step2:* The seam feature point is extracted and its 3D coordinate in the world coordinate system is determined. The $z$ coordinate of the feature point varies slowly when the laser line doesn't arrive at initial point. Thus, the initial welding point is determined at the position where the $z$ coordinate changes abruptly.

### B. Structured light methods based on phase shifting profilometry

The weld seam 3D measurement system based on PSP calculates the phase map of the weldment based on multiple phase-shifted sinusoidal patterns for high-resolution reconstruction of the 3D weld seam. Although PSP is the major measuring technique, other techniques may be incorporated to resolve the bottleneck problem of the phase unwrapping for PSP. For instance, both the sinusoidal patterns with shifted phases calculating the wrapped phase map and the gray code patterns helping to unwrap the wrapped phase map are used in [64]. These patterns are projected onto weldment one by one sequentially while the weldment is kept still. The wrapped phase map is computed analytically with at least 3 shifted phase values. After phase unwrapping, the absolute phase map is obtained and will be used to compute the 3D point cloud of the weldment based on the system parameters obtained by 3D calibration. The principle of computing the wrapped phase map by PSP is as follows.

The coded phase shifting sinusoidal patterns could be formulated as:

$$I_i(x, y) = 1 + \cos(2\pi f y + \alpha_i), i = 1, ..., N \quad (1)$$

where $(x, y)$ denotes the index of the sinusoidal pattern, $f$ is the carrier frequency and $\alpha_i$ is the $i$th shifted phase value. $N$ denotes the total number of the shifted phases. To cancel the reflectivity term, the minimum value required for $N$ is 3. In practical applications, the measurement accuracy when $N = 4$ is usually more accurate than the measurement accuracy when $N = 3$ [81]. Thus, $N$ is usually chosen as 4. For instance, 3 carrier frequencies and 4 phase shifting values are used in [40] to measure the weld bead. Suppose the shifted phase values are $0, \pi/2, \pi$ and $3\pi/2$ respectively, the wrapped phase map is obtained by solving the simultaneous equations [81].

$$2\pi f y + \varphi(x, y) = \arctan\left(\frac{\bar{I}_4(x, y) - \bar{I}_2(x, y)}{\bar{I}_1(x, y) - \bar{I}_3(x, y)}\right) \quad (2)$$

Since spatial phase unwrapping remained as a bottleneck problem for PSP to measure the complex shapes so far, the temporal phase unwrapping is usually used for phase unwrapping, which requires multiple carrier frequencies and more sinusoidal patterns as adopted in [40]. Similarly, the gray code could be used for phase unwrapping and only one high carrier frequency is needed as adopted in [64].

### C. Active stereo vision methods based on multiple line patterns

The weld seam 3D measurement system based on multiple line ASV calculates the disparity map of the weldment based on the multiple matched lines for medium-resolution reconstruction of the 3D weld seam [66-70]. Its working principle is very simple and could be summarized as follows.

*Step 1:* The parallel lines are distorted by the weldment and acquired by two stereo cameras.

*Step 2:* The distorted lines are extracted by image processing algorithms and matched between two camera views.

*Step 3:* All the matched lines are used to compute the disparity map that is used to reconstruct the 3D shape of the weldment with the system parameters obtained by 3D stereo calibration.

As can be seen, the above multiple line pattern based ASV method is single-shot. To reduce the image processing and stereo matching errors, the intervals between the adjacent lines must be large enough, which may decrease the reconstruction resolution and reconstruction accuracy significantly. A more accurate multiple line based ASV method is presented in [71-72] that utilizes a sequence of gray code patterns to generate the dense parallel lines. The gray code patterns are designed in a sparse to dense manner to facilitate the stereo matching process. The edge lines of the sparsest gray code pattern in two camera views are extracted and matched at first. The edge lines of the second sparest gray code pattern are extracted and matched based on the previously matched sparsest gray code pattern. So forth, the edge lines of other gray code patterns are matched. Compared to the single-shot method in [66], the ASV method based on multiple gray code patterns in [71-72] is more robust. However, the 3D measurement process is very inefficient. In addition, it may be difficult to keep the weldment still in the welding environments while the gray code patterns are projected on the weldment one by one sequentially. compromise method is to design the RGB line pattern in a coarse-to-fine (sparse-to-dense) manner [81]. Accordingly, the 3D shape can not only be measured efficiently by single-shot, but also be measured with increased resolution and accuracy compared to the method in [66].

### D. Active stereo vision methods based on Fourier transform profilometry

The weld seam 3D measurement system based on FTP ASV calculates the disparity map of the weldment based on the matched phase map for high-resolution reconstruction of the 3D weld seam [73]. Its working principle could be summarized as follows.

*Step 1:* One single high carrier frequency sinusoidal pattern is projected on the weldment. The wrapped phase map is computed by FTP and is then unwrapped with spatial phase unwrapping algorithms.

*Step 2:* The unwrapped phase map is matched between camera views to compute the disparity map that is used to reconstruct the 3D shape of the weldment with the system parameters obtained by 3D stereo calibration.

The principle of computing the wrapped phase map by FTP is as follows.

Firstly, The sinusoidal pattern is coded without phase shifting, i.e. $\alpha_i = 0$:

$$I(x, y) = 1 + \cos(2\pi f y) \quad (3)$$

The pattern image acquired by the CCD camera is formulated as:

$$g(x, y) = \gamma(x, y)(1 + \cos(2\pi f y + \varphi(x, y))) \quad (4)$$

where $\gamma(x, y)$ is the non-uniform reflectivity distribution and $\varphi(x, y)$ is the phase caused by the depth. The wrapped phase is obtained based on Fourier transform [81].

$$2\pi f y + \varphi(x, y) = \text{Im}\{\log[g^H(x, y)]\} \quad (5)$$

where $g^H(x, y)$ denotes the high frequency part of $g(x, y)$. After phase unwrapping, the absolute phase map is obtained and it can be used to compute the 3D point cloud of the object alone based on the system parameters obtained by 3D calibration. Only the spatial phase unwrapping algorithms could be used to unwrap the wrapped phase map for FTP, which remained as a bottleneck problem for FTP to measure the complex shapes so far. Fortunately, the shapes of most weldments are simple, i.e. lack of severe occlusions and large discontinuities. For FTP ASV, the unwrapped phase maps are matched to obtain the disparity map [73]. The 3D point cloud of the weldment is then reconstructed with the disparity map and the system parameters obtained by 3D calibration. Because of the accumulation of the FTP phase error and the stereo matching error, the accuracy of the proposed system in [73] will be inferior to that of FTP alone or ASV alone on the premise of the same hardware.

### E. Active stereo vision methods based on laser speckle patterns

The weld seam 3D measurement system based on laser speckle ASV calculates the disparity map of the weldment based on the matched laser speckle images for medium-resolution reconstruction of the 3D weld seam. The RealSense D435 camera has been widely used in robotic welding path planning [74-79] and is one of the most representative ASV methods that utilize the laser speckle patterns. The principle of the laser speckle based ASV is as follows.

*Step 1:* The randomly generated laser speckle patterns are projected onto the measured object to add features to the texture-less regions of the measured object. The pattern images are acquired by two stereo cameras synchronously.

*Step 2:* The two pattern images are matched by the local methods or by the global methods. The typical local methods include block matching, gradient matching and feature matching. The typical global methods include graph-cut, belief propagation and dynamic programming. The working principle of the local methods is to search for the minimum error over local regions using a match metric [81]. The working principle of the global method is to find a disparity function that minimizes a global energy function [81].

*Step 3:* The matched pattern images are used to compute the disparity map that is used to reconstruct the 3D shape of the weldment with the system parameters obtained by 3D stereo calibration.

## VI. TUTORIAL OF THE 3D CALIBRATION METHODS FOR THE ACTIVE VISUAL SENSING SYSTEMS

### A. The 3D calibration method for structured light systems

As reviewed in the above sections, the active visual sensing systems based on SL acquired the SL patterns with a camera and then calculate the seam tracking feature points [9-32], or detect the weld bead defects [33-40], or measure the weld seams [54-65] in the camera coordinate system. However, the welding robots operate the welding torch in the world coordinate system. Thus, 3D calibration is required to transform the camera coordinate system of the SL based active visual sensing system into the world coordinate system. However, few of the reviewed literatures address this important issue in details. Some literatures give the vision model of 3D reconstruction [33-34]. However, the given vision model is too simple to compute the 3D world coordinates of the weld seam, the weldment and the welding torch robustly.

Most importantly, the weld pool surface is mirror reflection instead of diffuse reflection. Thus, the 3D calibration of the weld pool geometry measurement system is different from those of diffuse surface measurement systems [9-32, 33-40, 54-65]. However, most literatures [42-48] about weld pool geometry measurement omit this important issue. Here, we give a tutorial of the 3D calibration method for the SL based active visual sensing systems in intelligent welding robots and it combines two parts: (1) the weld pool mirror surface measurement system 3D calibration and (2) the weldment diffuse surface measurement system 3D calibration. Independently, part (1) could be used for weld pool geometry measurement systems [42-48] and part (2) could be used for seam tracking systems [9-32], weld bead defection systems [33-40] and weld seam 3D measurement systems [54-65].

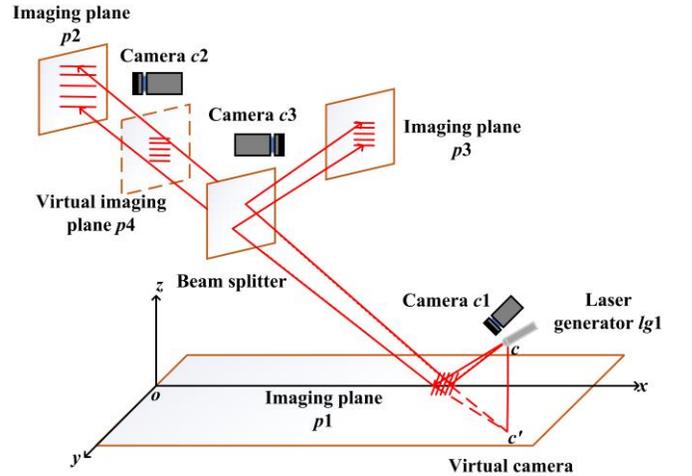

Fig. 5. The schematic diagram of the 3D calibration method for the weld pool mirror surface measurement system.

The weld pool mirror surface measurement system is calibrated as follows. As shown in Fig. 6, there are three cameras, i.e., $c1$, $c2$ and $c3$ that are aimed at three diffuse reflection planes $p1$, $p2$ and $p3$ respectively. The horizontal plane $p1$ at $z = 0$ is defined as the reference plane in the 3D world coordinate system and its origin is denoted as $o$. The parallel laser lines are projected onto $p1$ by a laser generator $lg1$ whose projection center is represented as $c$. The mirror image of the laser generator $lg1$ in the reference plane forms a virtual camera whose projection center is represented as $c'$. Camera $c1$ is used to calibrate the diffuse reflection plane $p1$ and computes the equations of the incident rays in the world coordinate system.

Each projected laser line corresponds to a number of incoming rays, and the specific number is determined by the camera image resolution. The equations of the incident rays are calculated by moving the horizontal diffuse reflection plane $p1$ at $z=-3$, $z=-2$, $z=-1$, $z=0$, $z=1$, $z=2$ and $z=3$ respectively in the vertical direction with a metric lab jack. In theory, two points can determine a straight line. In practice, in order to reduce the impact of noise, seven points and the least squares fitting method are used to determine the equation of each incident ray. In addition, the least squares estimation method is used to calculate the projection center $c=(c_x,c_y,f)$ of all the incident rays to determine the projection center of the virtual camera $c'=(c_x,c_y,-f)$.

After the equations of the incident rays are determined, the original diffuse reflection plane $p1$ at $z=0$ is replaced with a mirror reflection plane $p_m$ to form a virtual camera with a projection center $c'=(c_x,c_y,-f)$. After the parallel laser lines are projected onto the mirror reflection plane $p_m$, they are reflected onto the beam splitter that splits the reflected rays equally. Half rays are reflected onto and imaged at the diffuse reflection plane $p3$. $p4$ is the virtual diffuse reflection plane that intersects the reflected rays for the first time. $p4$ and $p3$ are symmetric, and the plane where the beam splitter is located is their center of symmetry. Half rays pass though the beam splitter and imaged at the diffuse reflection plane $p2$. Accordingly, $p2$ intersects the reflected rays for the second time. The equation of the diffuse reflection plane $p2$ or $p4$ in the 3D world coordinate system is computed based on the equation of $p_m$ or $p1$ at $z=0$ and the homography between them as follows.

Both the image plane $p_m$ of the virtual camera and the diffuse reflection plane $p1$ are at $z=0$. Their equations are same and denoted as $\pi_1 = [0,0,1,0]^T$. The rotation matrix $R=[r1,r2,r3]$ and the translation vector $t=[t_x,t_y,t_z]^T$ define the affine between $p2$ and $p_m$ or $p1$, and they are computed as follows.

$$r1 = [r_0, r_3, r_6]^T = \frac{K^{-1}h1}{\|K^{-1}h1\|} \quad (6)$$

$$r2 = [r_1, r_4, r_7]^T = \frac{K^{-1}h2}{\|K^{-1}h1\|} \quad (7)$$

$$r3 = [r_2, r_5, r_8]^T = r1 \times r2 \quad (8)$$

$$t = [t_x, t_y, t_z]^T = \frac{K^{-1}h3}{\|K^{-1}h1\|} \quad (9)$$

where $K$ is the intrinsic matrix of the virtual camera and it is computed as:

$$K = \begin{bmatrix} f & 0 & c_x \\ 0 & f & c_y \\ 0 & 0 & 1 \end{bmatrix} \quad (10)$$

$H$ is the homography between $p2$ and $p_m$ or $p1$.

$$H = [h1,h2,h3] = \begin{bmatrix} h_0 & h_1 & h_2 \\ h_3 & h_4 & h_5 \\ h_6 & h_7 & h_8 \end{bmatrix} \quad (11)$$

To compute $H$, we need to use two sets of camera calibrated points from $p2$ and $p_m$ or $p1$ respectively.

Because $p_m$ is a mirror plane and the intersection points cannot be directly imaged on it, so the diffuse reflection plane $p1$ is used to intersect with the incident rays to get the first set of points that is denoted as $(x_1(i), y_1(i)), i=1,2,...,N$. Camera $c1$ is used to calibrated this point set. The point set from $p2$ is obtained as the set of all the intersection points of the reflected rays with the diffuse reflection plane $p2$ and it is denoted as $(x_2(i), y_2(i)), i=1,2,...,N$. Camera $c2$ is used to calibrated this point set. With the two calibrated point sets, a series of equations can be obtained as follows.

$$x_1(i) = \frac{h_0 x_2(i) + h_1 y_2(i) + h_2}{h_6 x_2(i) + h_7 y_2(i) + h_8}, i=1,2,...,N \quad (12)$$

$$y_1(i) = \frac{h_3 x_2(i) + h_4 y_2(i) + h_5}{h_6 x_2(i) + h_7 y_2(i) + h_8}, i=1,2,...,N \quad (13)$$

where $N$ is the total number of the incident rays. The above equation series are solved by singular value decomposition to obtain $H$. After $H$ is computed, the rotation matrix $R=[r1,r2,r3]$ and the translation vector $t=[t_x,t_y,t_z]^T$ are computed by Eqs. (15-18). The equation $\pi_2$ of $p2$ is then computed as:

$$\pi_2 = (P^T)^{-1} \pi_1 \quad (14)$$

$$P = \begin{bmatrix} 1 & 0 & 0 & -c_x \\ 0 & 1 & 0 & -c_y \\ 0 & 0 & 1 & f \\ 0 & 0 & 0 & 1 \end{bmatrix} \begin{bmatrix} r_0 & r_1 & r_2 & t_x \\ r_3 & r_4 & r_5 & t_y \\ r_6 & r_7 & r_8 & t_z \\ 0 & 0 & 0 & 1 \end{bmatrix} \quad (15)$$

With the known equations of $p2$, the 3D coordinates of the intersection points on it are determined as:

$$\begin{bmatrix} X_2(i) \\ Y_2(i) \\ 1 \end{bmatrix} = (H')^{-1} \begin{bmatrix} u_2(i) \\ v_2(i) \\ 1 \end{bmatrix} = \begin{bmatrix} h_0' & h_1' & h_2' \\ h_3' & h_4' & h_5' \\ h_6' & h_7' & h_8' \end{bmatrix}^{-1} \begin{bmatrix} u_2(i) \\ v_2(i) \\ 1 \end{bmatrix} \quad (16)$$

$$Z_2(i) = \frac{\pi_2(1)x_2(i) + \pi_2(2)y_2(i) + \pi_2(4)}{-\pi_2(3)} \quad (17)$$

where $(u_2(i), v_2(i))$ is the camera coordinate of the $i$th intersection point in the image plane of camera $c2$. $H'$ is the homography between $p2$ and the image plane of camera $c2$. To compute $H'$, we need to use two sets of points from the diffuse reflection plane $p2$ and camera $c2$ respectively. The set of points from $c2$ is $(u_2(i), v_2(i)), i=1,2,...,N$ and the set of points from $p2$ is computed as follows.

$$\begin{bmatrix} X_2(i) \\ Y_2(i) \\ Z_2(i) \\ 1 \end{bmatrix} = L(i)\pi_2, i=1,2,...,N \quad (18)$$

where $L(i)$ denotes the equations of the reflected rays:

$$L(i) = \begin{bmatrix} 0 & c_y x_1(i) - c_x y_1(i) & -fx_1(i) & x_1(i)-c_x \\ c_x y_1(i) - c_y x_1(i) & 0 & -fy_1(i) & y_1(i)-c_y \\ fx_1(i) & fy_1(i) & 0 & f \\ c_x - x_1(i) & c_y - y_1(i) & -f & 0 \end{bmatrix} \quad (19)$$

With the plane equation $\pi_2$ of $p2$, the camera coordinates $(u_2, v_2)$ of $c2$ and the homography $H'$ between $p2$ and the image plane of $c2$, the world coordinates of the intersection points on $p2$ are computed by Eq. (16) and Eq. (17). In the same way, the equation $\pi_4$ of the virtual imaging plane $p4$, the homography $H''$ between $p4$ and the image plane of camera $c3$ and the world coordinates of the intersection points on the virtual imaging plane $p4$ are computed. The equation of a reflected ray in the world coordinate system can be determined by the three-dimensional world coordinates of its two intersection points with $p2$ and $p4$ respectively. According to the known equations of the incident ray and its corresponding reflected ray, the intersection points on the mirror reflection surface can be calculated by analytical solution.

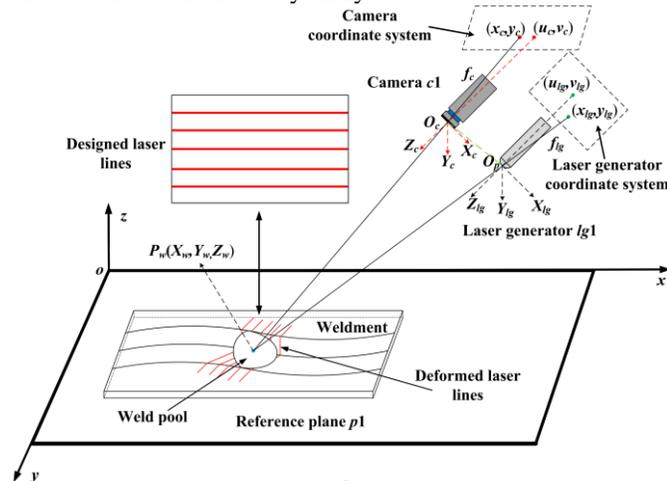

Fig. 6. The schematic diagram of the 3D calibration method for the weldment diffuse surface measurement system.

In order to monitor weld penetration [42-51] and weld bead defects [33-36] online, we need to know not only the three-dimensional shape of the weld pool, but also the three-dimensional shape information of the weldment around the weld pool. The weldment around the molten pool is a diffuse reflection surface and it needs the diffuse surface measurement calibration. The weldment diffuse surface measurement system is calibrated as follows.

In Fig. 6, the mathematical model between the camera coordinate system and the world coordinate system is formulated as:

$$Z_c \begin{bmatrix} x_c \\ y_c \\ 1 \end{bmatrix} = \begin{bmatrix} f_x & 0 & u_c \\ 0 & f_y & v_c \\ 0 & 0 & 1 \end{bmatrix} \begin{bmatrix} r_0 & r_1 & r_2 & t_x \\ r_3 & r_4 & r_5 & t_y \\ r_6 & r_7 & r_8 & t_z \end{bmatrix} \begin{bmatrix} X_w \\ Y_w \\ Z_w \\ 1 \end{bmatrix} \quad (20)$$

where $(x_c, y_c)$ is the camera coordinate and $(X_w, Y_w, Z_w)$ is the world coordinate. $f_c = (f_x, f_y)$ is camera's focal length and $(u_c, v_c)$ is the principle point of the camera. After matrix multiplication, Eq. (29) becomes the following format:

$$Z_c \begin{bmatrix} x_c \\ y_c \\ 1 \end{bmatrix} = \begin{bmatrix} m_{11}^{wc}, m_{12}^{wc}, m_{13}^{wc}, m_{14}^{wc} \\ m_{21}^{wc}, m_{22}^{wc}, m_{23}^{wc}, m_{24}^{wc} \\ m_{31}^{wc}, m_{32}^{wc}, m_{33}^{wc}, m_{34}^{wc} \end{bmatrix} \begin{bmatrix} X_w \\ Y_w \\ Z_w \\ 1 \end{bmatrix} \quad (21)$$

Let $m_{34}^{wc} = 1$ and write Eq. (30) in the form of simultaneous equations.

$$\begin{cases} Z_c x_c = m_{11}^{wc} X_w + m_{12}^{wc} Y_w + m_{13}^{wc} Z_w + m_{14}^{wc} \\ Z_c y_c = m_{21}^{wc} X_w + m_{22}^{wc} Y_w + m_{23}^{wc} Z_w + m_{24}^{wc} \\ Z_c = m_{31}^{wc} X_w + m_{32}^{wc} Y_w + m_{33}^{wc} Z_w + 1 \end{cases} \quad (22)$$

To separate parameters from the coefficients, the above simultaneous equations are then converted as:

$$\begin{bmatrix} X_w, Y_w, Z_w, 1, 0, 0, 0, 0, -x_c X_w, -x_c Y_w, -x_c Z_w \\ 0, 0, 0, 0, X_w, Y_w, Z_w, 1, -y_c X_w, -y_c Y_w, -y_c Z_w \end{bmatrix} \theta_c = \begin{bmatrix} x_c \\ y_c \end{bmatrix} \quad (23)$$

where

$$\theta_c = \begin{bmatrix} m_{11}^{wc}, m_{12}^{wc}, m_{13}^{wc}, m_{14}^{wc}, m_{21}^{wc}, m_{22}^{wc}, m_{23}^{wc}, m_{24}^{wc}, m_{31}^{wc}, m_{32}^{wc}, m_{33}^{wc} \end{bmatrix}^T$$

Let $X_c = \begin{bmatrix} X_w, Y_w, Z_w, 1, 0, 0, 0, 0, -x_c X_w, -x_c Y_w, -x_c Z_w \\ 0, 0, 0, 0, X_w, Y_w, Z_w, 1, -y_c X_w, -y_c Y_w, -y_c Z_w \end{bmatrix}$ and

$Y_c = \begin{bmatrix} x_c \\ y_c \end{bmatrix}$, we get the following equation.

$$X_c \theta_c = Y_c \quad (24)$$

The unknown parameter $\theta_c$ are solved by least squares estimation.

$$\theta_c = (X_c^T X_c)^{-1} X_c^T Y_c \quad (25)$$

In Fig. 6, the mathematical model between the projector coordinate system or the laser line generator coordinate system and the world coordinate system is formulated as:

$$Z_p \begin{bmatrix} x_p \\ y_p \\ 1 \end{bmatrix} = \begin{bmatrix} m_{11}^{wp}, m_{12}^{wp}, m_{13}^{wp}, m_{14}^{wp} \\ m_{21}^{wp}, m_{22}^{wp}, m_{23}^{wp}, m_{24}^{wp} \\ m_{31}^{wp}, m_{32}^{wp}, m_{33}^{wp}, m_{34}^{wp} \end{bmatrix} \begin{bmatrix} X_w \\ Y_w \\ Z_w \\ 1 \end{bmatrix} \quad (26)$$

where $(x_p, y_p)$ is the projector coordinate or the laser line generator coordinate. Since only $y_p$ is required for the 3D reconstruction, the following equation is obtained in the same way as described above.

$$X_p \theta_p = Y_p \quad (27)$$

where $X_p = \begin{bmatrix} X_w, Y_w, Z_w, 1, -y_p X_w, -y_p Y_w, -y_p Z_w \end{bmatrix}$,

$\theta_p = \begin{bmatrix} m_{21}^{wp}, m_{22}^{wp}, m_{23}^{wp}, m_{24}^{wp}, m_{31}^{wp}, m_{32}^{wp}, m_{33}^{wp} \end{bmatrix}^T$ and $Y_p = \begin{bmatrix} y_p \end{bmatrix}$.

The unknown parameter $\theta_p$ are solved by least squares estimation.

$$\theta_p = (X_p^T X_p)^{-1} X_p^T Y_p \qquad (28)$$

Before placing the weldment, the parallel laser lines are first projected onto the diffuse reflection plane $p1$. The laser lines are kept perpendicular to the $x$ axis. the axes $x$, $y$ and $z$ are determined during the 3D calibration process of the weld pool mirror surface measurement system described above, so as to ensure that the weld pool mirror surface measurement system and the weldment diffuse surface measurement system are in the same world coordinate system. Firstly, the metric lab jack is used to move the diffuse reflection plane $p1$ to the location $z = 0$, and the midpoint and two endpoints of each laser line are extracted by the camera $c1$ to obtain a set of known points, $(x_0(i), y_0(i), 0), i = 1, 2, ..., 15$. Here, we assume that there are five parallel laser lines. Secondly, the metric lab jack is then used to move the diffuse reflection plane $p1$ to the location $z = 0$, and the camera $c1$ extracts the midpoint and two endpoints of each laser line to obtain a set of known points $(x_1(i), y_1(i), 1), i = 1, 2, ..., 15$. With two sets of known points, $\theta_c$ and $\theta_p$ can be calculated by Eq. (25) and Eq. (28) respectively.

The 3D measurement process of the weldment around the weld pool is as follows:

*Step 1:* The parallel laser lines projected onto the weldment around the weld pool are distorted by the three-dimensional shape of the weldment and imaged in the camera $c1$. Firstly, the acquired laser line image is segmented. Secondly, the segmented laser lines are clustered from top to bottom. Finally, the center lines of all the cluster lines are extracted.

*Step 2:* The $y$ coordinate, $y_{lg}$ of any point on the clustered center line is defined as the phase of this point. The phases of all the points on a clustered center line form the phase vector of this clustered center line.

*Step 3:* With the known camera coordinate $x_c$, the phase vector index $y_c$ and the phase value $y_{lg}$ of any point on a clustered center line, the world coordinates of this point is computed by the following equation.

$$\begin{bmatrix} X_w \\ Y_w \\ Z_w \end{bmatrix} = H^{-1} \begin{bmatrix} x_c - m_{14}^{wc} \\ y_c - m_{24}^{wc} \\ y_p - m_{14}^{wp} \end{bmatrix} \qquad (29)$$

where $H = \begin{bmatrix} m_{11}^{wc} - x_c m_{31}^{wc}, m_{12}^{wc} - x_c m_{32}^{wc}, m_{13}^{wc} - x_c m_{33}^{wc} \\ m_{21}^{wc} - y_c m_{31}^{wc}, m_{22}^{wc} - y_c m_{32}^{wc}, m_{23}^{wc} - y_c m_{33}^{wc} \\ m_{11}^{wp} - y_p m_{21}^{wp}, m_{12}^{wp} - y_p m_{22}^{wp}, m_{13}^{wp} - y_p m_{23}^{wp} \end{bmatrix}$.

Since both the weldment diffuse surface measurement system and the weld pool mirror surface measurement system are calibrated by camera $c1$, their 3D measurement results are in the same world coordinate system. Therefore, the penetration state of the weld pool can be monitored online by comparing the height and width of the weld pool with the height and width of the surrounding weldment. The weld penetration monitoring process is illustrated in in Fig. 7. Fig. 7 (a) shows lack of penetration, where the height of the molten pool is normal and the width of the molten pool is rather small. Fig. 7 (b) shows complete penetration, where both the height and the width of the molten pool are normal. Fig. 7 (c) shows burn through, where the height of the molten pool is negative and the width of the molten pool is rather large.

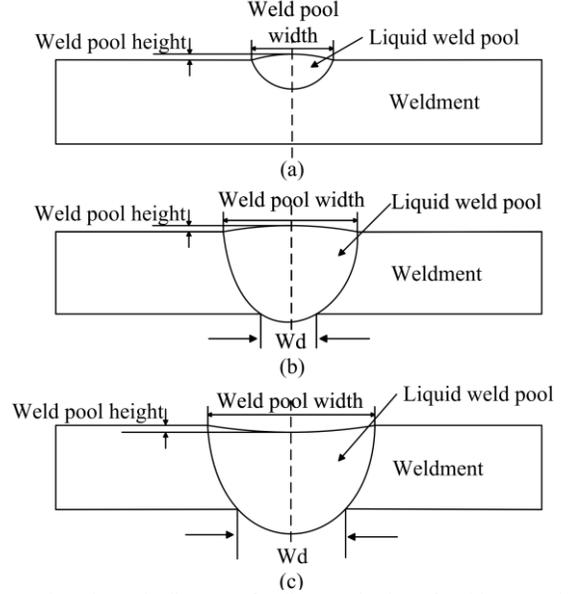

Fig. 7. The schematic diagram of online monitoring of weld penetration.

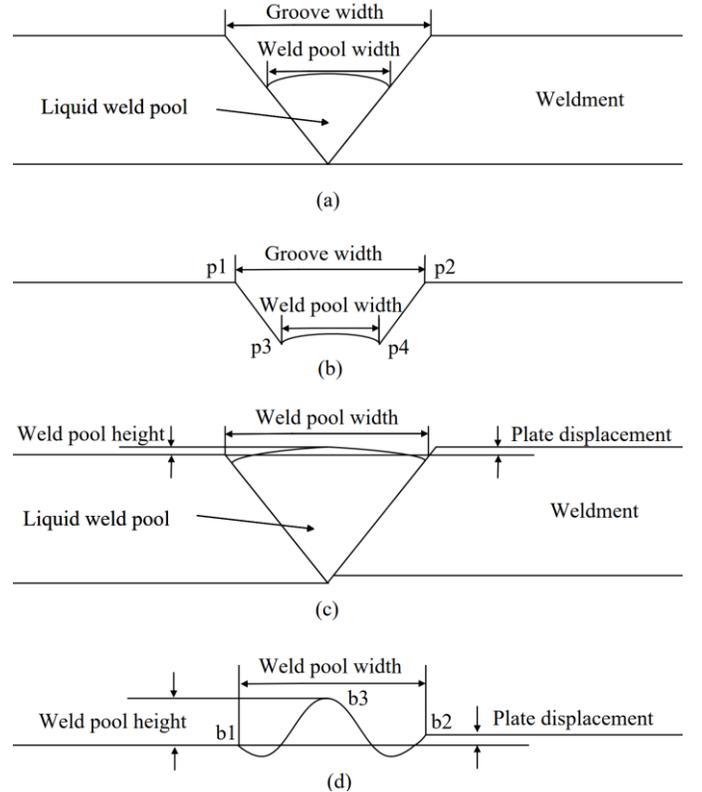

Fig. 8. The schematic diagram of online monitoring of weld bead defect.

Similarly, weld bead defects can be monitored online by comparing the height and width of the weld pool with the left and right height and groove width of the surrounding weldment. The online monitoring of weld bead defects is illustrated in Fig. 8. Fig. 8 (a-b) illustrates the monitoring process for weld misalignment defects when the torch deviates from the weld, which is detected according to the transverse symmetry of the weld groove. $p_1$, $p_2$, $p_3$ and $p_4$ are obtained by the feature point extraction method from the 3D shape of the weldment measured in real time. Fig. 8 (c-d) demonstrates the monitoring process of the undercut defect, the reinforcement height mutation defect and the plate displacement defect. The width of the weld pool determines the width of the weld bead, and when the width of the weld bead or the weld pool is greater than the width of the groove, the undercut defect will occur. The height of the weld pool determines the reinforcement height of the weld bead. When the height of the weld pool is too large or there is a large and sudden change, the weld reinforcement height defect will occur. The plate displacement defect occurs when the plate displacement of the weldment is greater than the acceptable threshold. These three types of defects could be detected with the feature points $b_1$, $b_2$ and $b_3$ in real time on line.

In summary, both the weld pool mirror surface measurement system calibration and the weldment diffuse surface measurement system calibration are essential for on-line weld penetration control [42-51] and the in-situ weld bead defect detection [33-36]. Only the weldment diffuse surface measurement system calibration is required for the SL based seam tracking systems [9-32] and the SL based weld seam 3D measurement systems [54-65].

### B. The 3D calibration method for active stereo vision systems

As reviewed in the above sections, the 3D measurement systems based on ASV calculate the 3D weld seams for off-line welding path planning [66-79]. Most researchers adopted the existing ASV products instead of designing their own ASV measurement systems. For instance, the chishine surface 120 product was used in [66-70] and the RealSense D435 was used in [74-79]. The major drawback of using the exiting 3D measurement products is that these products could only generate the point cloud of the whole weldment instead of generating the point cloud of the weld seam alone, which makes the measurement time increased enormously. In addition, specific weld seam extraction algorithms are required to extract the 3D seam from the reconstructed point cloud, which increases the computation time further. That explains why researchers could only use the measured 3D point cloud for offline welding path planning [66-79] instead of online welding path planning.

As a matter of fact, the 3D weld seam could be measured alone with the extracted weld seam feature points, which can increase the efficiency of welding path planning tremendously. To compute the 3D world coordinates of the feature points, ASV is the best choice because the matching between two cameras is more robust than the matching between the camera and the projector or laser generator used by SL. Here, we introduce a basic 3D calibration method that could be applied to all the ASV base active visual sensing systems reviewed in this paper.

The 3D model of the stereo vision is illustrated in Fig. 9, where $O$ and $O'$ denote the optical centers of the right camera and the left camera respectively. $o$ and $o'$ denote the principle point of the right camera and the principle point of the left camera respectively. The right angles include $\angle OoB = 90^o$, $\angle O'o'B' = 90^o$, $\angle ABC = 90^o$ and $\angle A'B'C' = 90^o$. $\pi$ denotes the image plane' equation of the right camera and $\pi'$ denotes the image plane' equation of the left camera. $b$ is the baseline between the left camera and the right camera, i.e. $OO' = b$. $Z = |PE|$ denotes the depth of the 3D world point $P$ and $\angle PEE' = 90^o$. $Z' = |PE'|$ denotes the distance of the 3D world point $P$ to the baseline. $\angle PE'O = \angle PE'O' = 90^o$. $p$ denotes the imaged point of the 3D world point $P$ in the right image plane. $p'$ denotes the imaged point of the 3D world point $P$ in the left image plane. The equations of the two cameras' image planes are formulated as $\pi = [0,0,-1,f]$ and $\pi' = [0,0,-1,f']$ respectively. The 3D coordinate of the world point could be computed as the 3D intersection point of two 3D rays. The two 3D rays are formulated by the following equations according to the stereo vision 3D model.

$$\begin{cases} \dfrac{x-O(1)}{(p(1)-O(1))} = \dfrac{y-O(3)}{(p(2)-O(2))} = \dfrac{z-O(3)}{(f-O(3))} = t \\ \dfrac{x-O'(1)}{(p'(1)-O(1))} = \dfrac{y-O'(3)}{(p'(2)-O'(2))} = \dfrac{z-O'(3)}{(f'-O'(3))} = t' \end{cases} \quad (30)$$

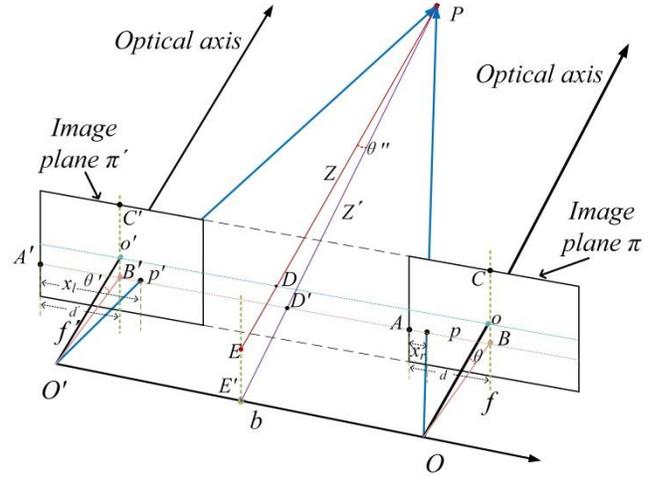

Fig. 9. The ray intersection based 3D model of stereo vision.

The 3D coordinate of the world point could be computed by solving the simultaneous equations in (39). The focal lengths and the optical centers of both cameras are determined by stereo vision calibration with a camera calibration board. $p$ and $p'$ are matched to compute the 3D coordinate of the world point.

The ray-intersection based 3D reconstruction method described above could be simplified into the triangulation based method as follows. The following assumptions need to be made: (1) the focal lengths of the two cameras are the same, i.e. $Oo = O'o' = f = f'$; (2) the image planes of the two cameras are the same, i.e. $|AB| = |A'B'| = d = d'$, $|CB| = |C'B'|$, and $|Co| = |C'o'|$; (3) the image planes of the two cameras are co-planar and the rows are in alignment, i.e. $\theta = \theta' = \theta''$ and $|OB| = |O'B'| = \dfrac{f}{\cos\theta}$. Then, the following equation is derived by the similar triangles, $OBp$ and $PD'p$.

$$\frac{PD'}{OB} = \frac{pD'}{Bp} \quad (31)$$

$$pD' = \frac{PD'(d - x_r)}{OB} = \frac{PD'(d - x_r)\cos\theta}{f} \quad (32)$$

The following equation is derived by the similar triangles, $O'B'p'$ and $PD'p'$.

$$\frac{PD'}{O'B'} = \frac{p'D'}{B'p'} = \frac{p'D'}{x_l - d} \quad (33)$$

$$p'D' = \frac{PD'(x_l - d)}{O'B'} = \frac{PD'(x_l - d)\cos\theta}{f} \quad (34)$$

The following equation is derived by the similar triangles, $PO'O$ and $Pp'p$.

$$\frac{PD'}{Z'} = \frac{pD' + p'D'}{b} \quad (35)$$

Substituting Equation (32) and (34) into Equation (35), we get the general equation to compute the depth $Z$ for the triangulation based method.

$$Z = Z'\cos\theta = \frac{fb}{(x_l - x_r)} \quad (36)$$

The focal length $f$ and the baseline $b$ between the two cameras are determined by stereo vision calibration with a camera calibration board. $x_r$ and $x_l$ are matched to compute the 3D coordinate of the feature point. Accordingly, the ray intersection based 3D model of stereo vision could be simplified to be the triangulation based 3D model as shown in Fig. 10. The weld seam feature point could be extracted robustly with some feature extraction algorithms. The world coordinates of all the extracted weld seam points are computed by Eq. (30) and then used to fit the 3D weld seam. As can be seen, the efficiency has been increased tremendously compared to those of exiting methods [66-79].

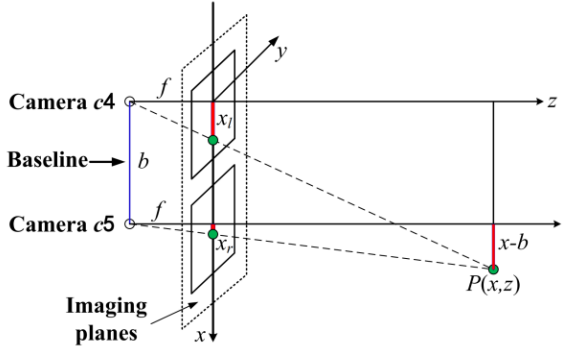

Fig. 10. The triangulation-based 3D model of stereo vision.

## VII. COMPARISON AND PROSPECT

In this section, we give a comprehensive comparison of all the reviewed active sensing methods for robotic welding in Table 1. The compared aspects include the required hardware, the cost, the number of the required projections, the efficiencies, the resolutions, the accuracies and the required key techniques. In the table, online seam tracking (ST) based on single laser line structured light (SLL-SL) [9-32] is denoted as online ST SLL-SL. The online weld bead defect detection (WBDD) based on single laser line structured light (SLL-SL) [33-36] is denoted as online WBDD SLL-SL. The offline weld bead defect detection (WBDD) based on single laser line structured light (SLL-SL) [37-38] is denoted as offline WBDD SLL-SL. The offline weld bead defect detection (WBDD) based on multiple laser line structured light (MLL-SL) [39] is denoted as offline WBDD MLL-SL. The offline weld bead defect detection (WBDD) based on phase shifting profilometry structured light (PSP-SL) [40] is denoted as offline WBDD PSP-SL. The online weld pool geometry measurement (WPGM) based on iterative estimation and laser dot matrix structured light (IE-LDM-SL) [42-44] is denoted as online WPGM IE-LDM-SL. The online weld pool geometry measurement (WPGM) based on iterative estimation and multiple laser line structured light (IE-MLL-SL) [45-48] is denoted as online WPGM IE-MLL-SL. The online weld pool geometry measurement (WPGM) based on ray intersection and laser dot matrix structured light (RI-LDM-SL) [50-51] is denoted as online WPGM RI-LDM-SL. The offline welding path planning (WPP) based on single laser line structured light (SLL-SL) [54-63] is denoted as offline WPP SLL-SL. The offline welding path planning (WPP) based on phase shifting profilometry structured light (PSP-SL) [64-65] is denoted as offline WPP PSP-SL. The offline welding path planning (WPP) based on multiple line active stereo vision (ML-ASV) [66-72] is denoted as offline WPP ML-ASV. The offline welding path planning (WPP) based on Fourier transform profilometry active stereo vision (FTP-ASV) [73] is denoted as offline WPP FTP-ASV. The offline welding path planning (WPP) based on laser speckle pattern active stereo vision (LSP-ASV) [74-79] is denoted as offline WPP LSP-ASV.

In addition, laser generator is denoted as LG, projector is denoted as P and camera is denoted as C. The structured light based weldment diffuse surface measurement system 3D calibration is denoted as SL-WDSMS-3DC, the structured light based weld pool mirror surface measurement system 3D calibration is denoted as SL-WPMSMS-3DC and the active stereo vision based weld seam diffuse surface measurement system 3D calibration is denoted as ASV-WSDSMS-3DC.

Based on the reviewed active visual sensing methods in this paper and their comprehensive comparisons in Table 1, we give the following prospects.

(1) The single laser line based SL method has been the most widely used method for seam tracking [9-32] in robotic welding so far because of its high accuracy, low cost and simplicity of implementation. The existing seam tracking methods [9-32] can only perform robust tracking for seams with simple shapes or single-layer and single-pass seams because they use only one tracking point. In the future, it may be combined with the online welding path planning to compute more tracking points for the seams with complex shapes or the multi-layer and multi-pass seams.

(2) The effectiveness of the single laser line based SL method for in situ weld bead defect detection has been verified in the past research [33-36]. However, when it detected considerable defects, it could only stop the welding process immediately for cost saving instead of adjusting the welding parameters to avoid or reduce the defects. In the future, a more promising way of in situ weld bead defect detection is to measure the 3D shape of the molten weld pool before it forms the bead since the 3D geometry of the weld pool determines the 3D geometry of the weld bead. When the 3D shape of the weld pool does not meet the conditions of forming the desired bead, the welding parameters are adjusted in situ to change the 3D geometry of the weld pool until it meets the conditions of the forming the desired bead.

(3) The advantage of the offline weld bead defect detection over the in-situ weld bead defect detection is that it has the potential to reconstruct the 3D weld bead as a whole more efficiently. However, the single laser line based offline welding bead defect detection [37-38] is very time consuming because the laser line needs to be moved to plenty of places for 3D reconstruction of the whole weld seam. On the contrary, the multiple laser line based SL method can increase the efficiency significantly while keeping the cost low [81]. In the future, the multiple laser lines that are as dense as possible should be used to reconstruct the 3D weld bead by single-shot.

**Table 1** Comprehensive comparison of the reviewed active visual sensing methods.

| Active sensing methods | Hardware | Cost | Projections | Efficiencies | Resolutions | Accuracies | Key techniques |
|---|---|---|---|---|---|---|---|
| online ST SLL-SL [9-32] | 1LG+1 C | Low | 1 | High | Low | High | SL-WDSMS-3DC |
| online WBDD SLL-SL [33-36] | 1LG+1 C | Low | 1 | High | Low | High | SL-WDSMS-3DC |
| offline WBDD SLL-SL [37-38] | 1LG+1 C | Low | ≥ 10 | Low | Low | High | SL-WDSMS-3DC |
| offline WBDD MLL-SL [39] | 1LG+1C | Low | ≥ 10 | Medium | Low | High | SL-WDSMS-3DC |
| offline WBDD PSP-SL [40] | 1P+1C | Low | ≥ 10 | Medium | High | High | SL-WDSMS-3DC |
| online WPGM IE-LDM-SL [42-44] | 1LG+1C | low | 1 | High | Low | Medium | Not available |
| online WPGM IE-MLL-SL [45-48] | 1LG+1C | Low | 1 | High | Medium | Medium | Not available |
| online WPGM RI-LDM-SL [50-51] | 1LG+2C | Low | 1 | High | Low | High | SL-WPMSMS-3DC |
| offline WPP SLL-SL [54-63] | 1LG+1 C | Low | ≥ 10 | Low | Low | High | SL-WDSMS-3DC |
| offline WPP PSP-SL [64-65] | 1P+1C | Low | ≥ 10 | Medium | High | High | SL-WDSMS-3DC |
| offline WPP ML-ASV [66-70] | 1P+2C | low | 1 | High | Medium | Medium | ASV-WSDSMS-3DC |
| offline WPP ML-ASV [71-72] | 1P+2C | low | ≥ 10 | Medium | Medium | High | ASV-WSDSMS-3DC |
| offline WPP FTP-ASV [73] | 1P+2C | Low | 1 | High | High | Medium | ASV-WSDSMS-3DC |
| offline WPP LSP-ASV [74-79] | 1P+2C | Low | 1 | High | Medium | Medium | ASV-WSDSMS-3DC |

(4) The PSP has been used to measure the weld bead for defect detection [40] and the weld seam for welding path planning [64-65]. However, only the time-consuming multiple carrier frequencies and multiple steps PSP methods have been used. In the future, real-time single high carrier frequency based PSP methods [80-81] should be adopted by robotic welding. Accordingly, the phase unwrapping problem can be addressed based on the shape characteristics of the weld bead or the weld seam.

(5) The FTP based ASV has been used to measure the weld seam for welding path planning [73]. However, the accumulation of the stereo matching error and FTP phase error makes its accuracy inferior to that of the FTP alone or that of the ASV alone [81]. In the future, FTP alone should be used to measure the weld bead or the weld seam because of its high accuracy, high efficiency and high resolution. The phase unwrapping problem can also be addressed based on the shape characteristics of the weld bead or the weld seam.

(6) The laser dot matrix SL based ray intersection method has been used to measure the 3D shape of the molten weld pool for penetration control [50-51]. However, it is very challenging to index the laser dots correctly for ray tracing when the shape of the weld pool changes rapidly. In addition, the resolution of the laser dot matrix is very limited compared to those of the multiple laser line patterns. In the future, the multiple laser line SL patterns based ray intersection method should be used to reconstruct the 3D shape of the molten weld pool online. The measured geometry of the weld pool not only can be used to monitor the weld penetration, but also can be used to monitor the weld bead defects. Consequently, the 3D calibration method in Section 6.1 will become very important to achieve the robust and simultaneous monitoring.

(7) Existing welding path planning techniques [54-79] are based on existing 3D measurement techniques (such as phase shift profilometry [64-65], ASV based on FTP [73]) or products (such as Chishine surface 120 [66-70]). RealSense D435[74-79]) to obtain the 3D point cloud of the entire weldment and then extract the 3D weld from the 3D point cloud, which is a time-consuming process, so it can only be used for offline weldment path planning. In the future, only the 3D seam should be reconstructed based on the 3D coordinates of the feature points for online welding path planning. Accordingly, the ASV will be the best choice to compute the 3D coordinates of the feature points because the matching between two cameras is more robust than the matching between the camera and the projector or laser generator used by SL [81]. Consequently, the 3D calibration method introduced in Section 6.2 will become very important to achieve robust online welding path planning.

(8) In the future, both online welding path planning and online weld pool monitoring will be indispensable for the next generation intelligent welding robots. Fusion of several active visual sensing methods in one welding robot will be a new trend. For instance, a multiple laser line SL based online welding path planning system and a multiple laser line based weld pool measurement system are fused to guide the movement of the welding torch, detect the weld bead defect and control the weld penetration simultaneously. Accordingly, the active visual sensing technology for robotic welding will be improved to a new level.

## VIII. CONCLUSION

Intending to present a valuable reference for researchers engaged in the related fields, we give a comprehensive review of recent advances in the active visual sensing methods for robotic welding in this study. The reviewed methods are classified into four categories based on their uses. Firstly, their working principles are described briefly. Secondly, the tutorials of 3D calibration for these active sensing methods that are missed in the reviewed literatures are given to fill the gaps, which are very important for the active visual sensing systems in the next generation intelligent robots. By calibrating the weld pool mirror surface SL measurement system and the weldment diffuse surface SL measurement system in the same world coordinate system, the weld penetration and the weld bead defect could be monitored simultaneously online. By measuring the 3D weld seam alone with the extracted weld seam feature points and ASV, the welding path could be planned online. Thirdly, prospects are given based on the comparisons of these active visual sensing methods with regard to their advantages and disadvantages.